\documentclass[conference]{IEEEtran}
\usepackage[utf8]{inputenc}

\usepackage[numbers]{natbib}

\usepackage{amsmath}
\usepackage[bb=dsserif]{mathalpha}
\usepackage{siunitx}
\usepackage{pifont}   

\usepackage{stmaryrd}   
\usepackage{trimclip}   

\usepackage{booktabs}
\usepackage{makecell}  
\usepackage[flushleft]{threeparttable}  
\usepackage{multirow}
\usepackage{multicol}
\usepackage{rotating}
\usepackage{adjustbox}

\usepackage[normalem]{ulem}

\usepackage{xspace}    
\usepackage{xcolor}    
\usepackage{colortbl}
\usepackage{soul}      
\usepackage{textcomp}  
\usepackage{gensymb}   
\usepackage{graphicx} 
\usepackage[linesnumbered,ruled,vlined]{algorithm2e}  
\usepackage{flushend}
\usepackage[bookmarks=true]{hyperref}

\renewcommand{\thefootnote}{\fnsymbol{footnote}}

\DeclareMathOperator*{\argmax}{arg\,max}

\begin{document}

\title{CoDEPS: Online Continual Learning for \\ Depth Estimation and Panoptic Segmentation}


\author{\authorblockN{Niclas Vödisch\textsuperscript{\footnotesize 1*},
Kürsat Petek\textsuperscript{\footnotesize 1*},
Wolfram Burgard\textsuperscript{\footnotesize 2}, and
Abhinav Valada\textsuperscript{\footnotesize 1}} \vspace{0.1cm}
\authorblockA{\textsuperscript{\footnotesize 1}University of Freiburg \quad\quad {\textsuperscript{\footnotesize 2}University of Technology Nuremberg}}
}


\makeatletter
\DeclareRobustCommand{\shortto}{%
  \mathrel{\mathpalette\short@to\relax}%
}
\newcommand{\short@to}[2]{%
  \mkern2mu
  \clipbox{{.25\width} 0 0 0}{$\m@th#1\vphantom{+}{\shortrightarrow}$}%
  }
\makeatother

\newcommand\blfootnote[1]{%
  \begingroup
  \renewcommand\thefootnote{}\footnote{#1}%
  \addtocounter{footnote}{-1}%
  \endgroup
}

\newcommand{\cmark}{\ding{51}}  
\newcommand{\cross}{\ding{61}}
\newcommand{\asterisk}{\ding{83}}

\newcommand{\blue}[1]{{\color{blue} #1}}
\newcommand{\red}[1]{{\color{red} #1}}
\definecolor{Gray}{gray}{0.9}

\newcommand{\refeqn}[1]{Eq.~\ref{#1}}
\newcommand{\refEqn}[1]{Equation~\ref{#1}}
\newcommand{\reffig}[1]{Fig.~\ref{#1}}
\newcommand{\refsec}[1]{Sec.~\ref{#1}}
\newcommand{\refalg}[1]{Algorithm~\ref{#1}}
\newcommand{\reftab}[1]{Table~\ref{#1}}

\newcolumntype{R}[2]{%
    >{\adjustbox{angle=#1,lap=\width-(#2)}\bgroup}%
    l%
    <{\egroup}%
}
\newcommand{\rot}[1]{\rotatebox[origin=c]{70}{#1}}

\newcommand{\net}{CoDEPS\xspace}


\maketitle

\begin{abstract}
    Operating a robot in the open world requires a high level of robustness with respect to previously unseen environments. Optimally, the robot is able to adapt by itself to new conditions without human supervision, e.g., automatically adjusting its perception system to changing lighting conditions.
In this work, we address the task of continual learning for deep learning-based monocular depth estimation and panoptic segmentation in new environments in an online manner. We introduce \net to perform continual learning involving multiple real-world domains while mitigating catastrophic forgetting by leveraging experience replay.
In particular, we propose a novel domain-mixing strategy to generate pseudo-labels to adapt panoptic segmentation.
Furthermore, we explicitly address the limited storage capacity of robotic systems by leveraging sampling strategies for constructing a fixed-size replay buffer based on rare semantic class sampling and image diversity.
We perform extensive evaluations of \net on various real-world datasets demonstrating that it successfully adapts to unseen environments without sacrificing performance on previous domains while achieving state-of-the-art results. The code of our work is publicly available at \mbox{\url{http://codeps.cs.uni-freiburg.de}}.

    \blfootnote{\textsuperscript{\footnotesize *}Equal contribution.}
\end{abstract}

\IEEEpeerreviewmaketitle

\section{Introduction}

Deploying robots such as autonomous cars in urban scenarios requires a holistic understanding of the environment with a unified perception of semantics, instances, and depth. The joint solution of these tasks enables vision-based methods to generate a 3D semantic reconstruction of the scene, which can be leveraged for downstream applications such as localization or planning. While deep learning-based state-of-the-art approaches perform well when inference is done under similar conditions as used for training, their performance can drastically decrease when the new target domain differs from the source domain, e.g., due to environmental conditions~\cite{valada2016towards}, different sensor parameters~\cite{bevsic2022unsupervised, cheng2022vsseg}. This domain gap poses a great challenge for robotic platforms that are deployed in the open world without prior knowledge about the target domain. Additionally, unlike the source domain where ground truth annotations are generally assumed to be known and can be used for the initial training, such supervision is not applicable to the target domain due to the absence of labels, rendering classical domain adaptation methods unsuitable. Unsupervised domain adaptation attempts to overcome these limitations. However, the vast majority of proposed approaches focuses on sim-to-real domain adaptation mostly in an offline manner~\cite{guizilini2021geometric, lopez2020desc}, i.e., a directed knowledge transfer without the need to avoid catastrophic forgetting and with access to abundant target annotations. Additionally, such works might not consider limitations on a robotic platform, e.g., available compute hardware and limited storage capacity~\cite{kuznietsov2021comoda, voedisch2023continual}.

\begin{figure}[t]
    \centering
    \includegraphics[width=\linewidth]{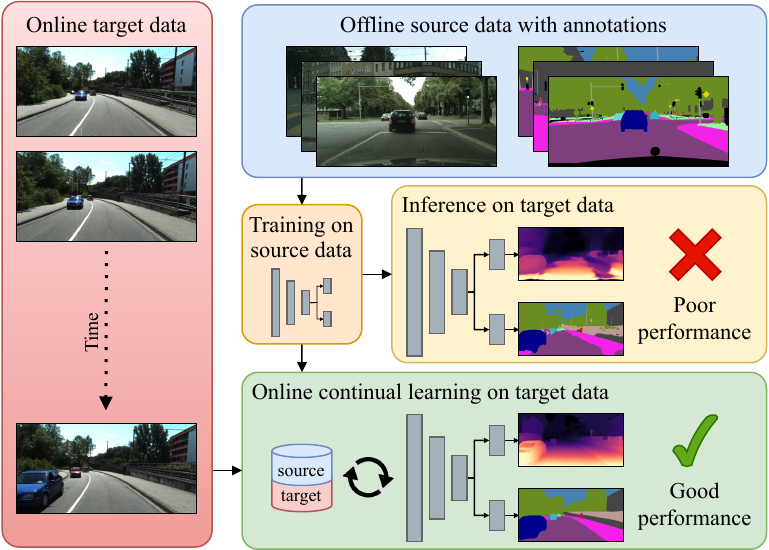}
    \vspace*{-.5cm}
    \caption{Neural networks often perform poorly when deployed on a target domain that differs from the source domain used for training. To close this domain gap, we propose to continuously adapt the network by exploiting online target images. To mitigate catastrophic forgetting and enhance generalizability, we leverage a fixed-size replay buffer allowing the method to revisit data from both the source and target domains.
    }
    \label{fig:teaser}
    \vspace*{-.5cm}
\end{figure}

In this work, we use online continual learning to address these challenges for depth estimation and panoptic segmentation in a multi-task setup. As shown in \reffig{fig:teaser}, we leverage images from an onboard camera to perform online continual learning enhancing performance during inference time. While a naive approach would result in overfitting to the current scene, our method \net mitigates forgetting by leveraging experience replay of both source data and previously seen target images. We combine a classical replay buffer with generative replay in the form of a novel cross-domain mixing strategy allowing us to exploit supervised ideas also for unlabeled target data. We explicitly address the aforementioned hardware limitations by using only a single GPU and restricting the replay buffer to a fixed size.
We demonstrate that \net successfully improves on new target domains without sacrificing performance on previous domains.

The main contributions of this work are as follows:
\begin{enumerate}
    \item We introduce \net, the first online continual learning approach for joint monocular depth estimation and panoptic segmentation.
    \item We propose a novel cross-domain mixing strategy to adapt panoptic segmentation to unlabeled target data.
    \item To address the storage restrictions of robotic platforms, we leverage a fixed-size replay buffer based on rare class sampling and image diversity.
    \item We extensively evaluate \net and compare it to other methods in challenging real-to-real settings.
    \item We release our code and the trained models at \mbox{\url{http://codeps.cs.uni-freiburg.de}}.
\end{enumerate}

\section{Related Work}
In this section, we provide an overview of monocular depth estimation, panoptic segmentation, and unsupervised domain adaptation including continual learning.


{\parskip=5pt
\noindent\textit{Monocular Depth Estimation:}
Monocular depth estimation is the task of predicting a dense depth map from a single RGB image. While supervised approaches exploit measurements from range sensors to supervise the network predictions~\cite{qiao2021vipdeeplab}, unsupervised methods leverage geometric cues from temporal context~\cite{godard2017unsupervised, zhou2017unsupervised}. Most of the research on unsupervised learning tackles the limitations of the so-called photometric loss function that is usually employed for unsupervised depth learning, e.g., dynamic object handling~\cite{bevsic2022dynamic, casser2019dynamic, li2021dynamic}, occlusion~\cite{godard2019digging}, and abrupt illumination changes~\cite{yang2020d3vo}.
In this work, we leverage Monodepth2~\cite{godard2019digging} for unsupervised depth learning and employ it similarly to \citet{guizilini2021geometric} for the purpose of domain adaptation.
}


{\parskip=5pt
\noindent\textit{Panoptic Segmentation:}
Panoptic segmentation unifies the two tasks of semantic and instance segmentation by fusing the respective targets into a joint output. Furthermore, semantic classes are grouped into ``stuff'' classes, e.g., \textit{road} or \textit{building}, and ``thing'' classes, e.g., \textit{car} or \textit{pedestrian}. In particular, the goal of vision-based panoptic segmentation is to assign a semantic class to every pixel of an image and an additional instance label to each object belonging to the ``thing'' classes. Panoptic segmentation networks usually comprise a joint encoder and separate decoders for each subtask, whose outputs are subsequently merged by a panoptic fusion module.
Existing works can be categorized into bottom-up~\cite{cheng2020panoptic, mohan2022perceiving} and top-down~\cite{gosala2022bird, mohan2022amodal} approaches. Whereas bottom-up methods detect instances in a proposal-free manner from the semantic prediction, top-down methods include an additional proposal generation step. Contradictions to the semantic predictions are then resolved during post-processing.
In this work, we build upon the bottom-up Panoptic-Deeplab~\cite{cheng2020panoptic} with changes to the semantic head according to~\citet{guizilini2021geometric}.
}


{\parskip=5pt
\noindent\textit{Unsupervised Domain Adaptation:}
Domain adaptation aims to bridge the domain gap between a source domain $\mathcal{S}$ used for training and a target domain $\mathcal{T}$ used for inference to mitigate a loss in performance.
An important aspect is whether the performance on the source domain must be maintained, linking domain adaptation to continual learning~(CL)~\cite{lopez2017gradient}, where the objective of a task or the task itself can change over time. A CL system has to adapt to the new target objective while retaining the knowledge to solve the previous task(s), i.e., avoiding catastrophic forgetting. Ideally, the CL system can further achieve positive forward transfer, i.e., improve on future yet untrained tasks given the current task.
In many real-world scenarios ground truth annotations for the target domain are not available, thus requiring unsupervised domain adaptation (UDA) methodology. Offline UDA assumes that abundant target data is accessible. However, in order to guarantee the continuous operation of a robot in new domains, UDA approaches have to work online without previous target data collection.}


Offline UDA can leverage both annotated source data and abundant unlabeled target data, enabling learning a given task from $\mathcal{S}$ while simultaneously adapting the network to $\mathcal{T}$.
For depth estimation, DESC~\cite{lopez2020desc} adapts from a synthetic source domain containing RGB images and ground truth depth to a real-world target domain by performing source-to-target style transfer and using a consistency loss between depth predictions from RGB and semantic maps.
GUDA~\cite{guizilini2021geometric} tackles UDA for semantic segmentation using depth estimation as a proxy task. A shared encoder with task-specific heads for depth estimation and semantic segmentation is trained via source supervision. Simultaneously, data from $\mathcal{T}$ is used to update the encoder and depth head in an unsupervised manner. Due to the refined weights of the encoder, the semantic predictions on $\mathcal{T}$ improve as well.
Another common approach for adapting semantic segmentation is cross-domain sampling enabling partial supervision on $\mathcal{T}$. DACS~\cite{tranheden2021dacs} mixes images from $\mathcal{S}$ and $\mathcal{T}$ by copying the pixels of a source image to a target image based on the semantic labels~\cite{olsson2021classmix}. The semantic prediction of the target image is updated with ground truth source labels for the same set of pixels. The network is then jointly trained on annotated source data and the pseudo-labeled mixing data.
Recently, ConfMix~\cite{mattolin2023confmix} proposed a simple yet effective mixing strategy for object detection, where a target image is divided into rectangular image regions. The region with the most confident predictions is then copied onto a source image and the respective ground truth annotations.
Finally, \citet{huang2021cross} propose a UDA method for panoptic segmentation by regularizing complementary features from semantic and instance segmentation.
In this work, we extend the aforementioned mixing strategies to instance-based sampling and explicitly address differing camera parameters.


During online UDA, samples from $\mathcal{T}$ can only be accessed in a consecutive manner resembling the image stream of a camera. Typically, a network is trained offline via supervision on $\mathcal{S}$ and then adapted to $\mathcal{T}$ during inference time. Such a setup rises two main challenges: first, incoming target samples originate from highly similar scenes and thus drastically reduce the diversity; second, this similarity of consecutive samples leads to a strong overfitting of the model to the scene~\cite{zhang2020online}.
Initial works for online UDA focused on depth learning~\cite{kuznietsov2021comoda, zhang2020online} and visual odometry~\cite{li2020self, voedisch2023continual}, for which unsupervised training schemes are already well established.
Whereas \citet{zhang2020online} propose novel network modules that are adapted via a meta-learning paradigm to mitigate forgetting, CoMoDA~\cite{kuznietsov2021comoda} employs a common CL strategy, i.e., experience replay to combine the online target sample with previously seen samples. Continual SLAM~\cite{voedisch2023continual} also uses unsupervised depth estimation as a proxy task to enhance visual odometry during inference time. Additionally, it demonstrates that incorporating samples from $\mathcal{S}$ and previous target domains $\mathcal{T}_i$ prevents catastrophic forgetting when revisiting domains.
Similar settings involving multiple target domains, which are hence closely related to classical CL, are also addressed for semantic segmentation. CBNA~\cite{klingner2022continual} mixes statistics from $\mathcal{S}$ and $\mathcal{T}$ to update the batch normalization layers and showcases the efficacy of the approach on continually visited target domains.
CoTTA~\cite{wang2022cotta} adapts the entire network without using source data but self-supervision. To tackle error accumulation, it uses an exponential moving average filter and student-teacher consistency when updating the network weights.
Using depth estimation as a proxy task, \citet{kuznietsov2022towards} extend GUDA~\cite{guizilini2021geometric} to online UDA with experience replay and confidence regularization on the semantic predictions.
To the best of our knowledge, we propose the first approach for online continual UDA for joint depth estimation and panoptic segmentation.

\section{Technical Approach}

The setting investigated in this work consists of two steps. First, we train a neural network on the source domain~$\mathcal{S}$ partly using ground truth supervision. Second, to close the gap between domains, we continuously adapt the network during inference time on the target domain~$\mathcal{T}$ using a replay buffer and unsupervised training strategies.

\begin{figure*}[t]
    \centering
    \includegraphics[width=\linewidth]{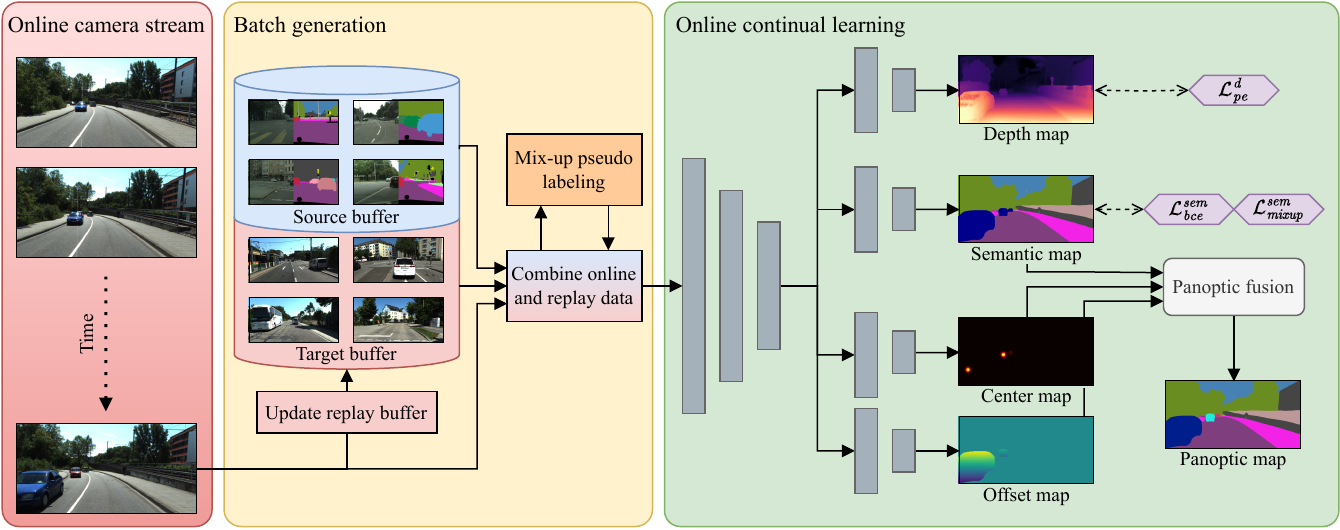}
    \vspace*{-.5cm}
    \caption{
    Overview of our proposed \net. Unlabeled RGB images from an online camera stream are combined with samples from a replay buffer comprising both annotated source samples and previously seen target images. Cross-domain mixing enables pseudo-supervision on the target domain. The network weights are then updated via backpropagation using the constructed data batch. The additional PoseNet required for unsupervised monocular depth estimation is omitted in this visualization.}
    \label{fig:overview}
    \vspace*{-.5cm}
\end{figure*}


\subsection{Network Architecture and Source Domain Pretraining}
\label{ssec:ta-network-pretraining}

In this section, we detail the network architecture and loss functions that we employ during the pretraining stage on the source domain.

{\parskip=5pt
\noindent\textit{Architecture:}
We build our network following a common multi-task design scheme, i.e., using a single backbone followed by task-specific heads. A high-level overview of the network architecture is shown in \reffig{fig:overview}. In detail, we use a ResNet-101~\cite{he2016deep} as the shared encoder for all three tasks including depth prediction, semantic segmentation, and instance segmentation. The depth head follows the design of Monodepth2~\cite{godard2019digging} comprising five consecutive convolutional layers with skip connections to the encoder. Additionally, we include a separate PoseNet consisting of a ResNet-18 encoder and a four-layer CNN to estimate the camera motion between two image frames. For panoptic segmentation, we follow the bottom-up method Panoptic-Deeplab~\cite{cheng2020panoptic}, leveraging separate heads for semantic segmentation and instance segmentation, and slightly modify the semantic head~\cite{guizilini2021geometric}. Specifically, the instance head consists of two sub-heads to predict the center of an object and to associate each pixel of an image to the corresponding object or the background. Finally, a panoptic fusion module~\cite{cheng2020panoptic} assigns a semantic label to the class-agnostic instance predictions using majority voting over the semantic predictions of all pixels within an instance.
}


{\parskip=5pt
\noindent\textit{Source Domain Pretraining:}
During the initial training phase on the source domain, we assume to have access to image sequences as well as ground truth panoptic segmentation annotations. In the following, we briefly describe the respective loss functions that we employ for training the three task-specific heads.
}

We train the depth estimation head using the common methodology of unsupervised training based on the photometric error~\cite{godard2019digging}. In particular, we leverage an image triplet $\{\mathbf{I_{t_0}}, \mathbf{I_{t_1}}, \mathbf{I_{t_2}}\}$ to predict depth $\mathbf{D_{t_1}}$ and camera motion $\mathbf{M_{t_0 \shortto t_1}}$ and  $\mathbf{M_{t_1 \shortto t_2}}$. Afterwards, we compute the photometric error loss $\mathcal{L}^d_\mathit{pe}$ as a weighted sum of the reprojection loss~$\mathcal{L}^d_\mathit{pr}$ and the image smoothness loss~$\mathcal{L}^d_\mathit{sm}$:
\begin{equation}
    \mathcal{L}^d_\mathit{pe} = \lambda_\mathit{pr} \mathcal{L}^d_\mathit{pr} + \lambda_\mathit{sm} \mathcal{L}^d_\mathit{sm}.
    \label{eqn:loss-photometric}
\end{equation}

We train the semantic segmentation head in a supervised manner using the bootstrapped cross-entropy loss with hard pixel mining~$\mathcal{L}^\mathit{sem}_\mathit{bce}$ following Panoptic-Deeplab~\cite{cheng2020panoptic}.

For training the instance segmentation head, we adopt the MSE loss~$\mathcal{L}^\mathit{ins}_\mathit{center}$ for the center head and the L1 loss~$\mathcal{L}^\mathit{ins}_\mathit{offset}$ for the offset head. The total loss to supervise instance segmentation is then computed as a weighted sum:
\begin{equation}
    \mathcal{L}^\mathit{ins}_\mathit{co} = \lambda_\mathit{center} \mathcal{L}^\mathit{ins}_\mathit{center} + \lambda_\mathit{offset} \mathcal{L}^\mathit{ins}_\mathit{offset}.
    \label{eqn:supervised-instance}
\end{equation}


\subsection{Online Adaptation}
\label{ssec:ta-online-adaptation}

After the described network has been trained on the source domain $\mathcal{S}$ using the aforementioned losses, we aim to adapt it to the target domain $\mathcal{T}$ in a continuous manner. That is, unlike other works, data from the target domain is revealed frame by frame resembling the online stream of an onboard camera.
As depicted in \reffig{fig:overview}, every adaptation iteration consists of the following steps:
\begin{enumerate}
    \item Construct an update batch by combining online and replay data.
    \item Generate pseudo-labels using the proposed cross-domain mixing strategy.
    \item Perform backpropagation to update the network weights.
    \item Update the replay buffer.
\end{enumerate}
In this section, we first detail the structure of the utilized replay buffer and then propose adaptation schemes for both depth estimation and panoptic segmentation.


{\parskip=5pt
\noindent\textit{Replay Buffer and Batch Generation:}
Upon receiving a new image taken by the robot's onboard camera, we construct a batch that is used to perform backpropagation on the network weights. In detail, a batch $\mathbf{b_t}$ consists of the current online image $\mathbf{I_t} \in \mathcal{T}$, previously received target images $\mathbf{I_{\mathcal{T}_i}} \in \mathbf{B_\mathcal{T}}$, and fully annotated source samples $\mathbf{I_{\mathcal{S}_i}} \in \mathbf{B_\mathcal{S}}$. Here, $\mathbf{B_\mathcal{T}} \subseteq \mathcal{T}$ and $\mathbf{B_\mathcal{S}} \subseteq \mathcal{S}$ denote the respective replay buffers.
Formally\footnote{To improve readability, we omit in the notation that each image sample includes its two previous frames enabling unsupervised depth estimation.},
\begin{equation}
    \mathbf{b_t} = \{\mathbf{I_{t}}, \mathbf{I_{\mathcal{T}_0}}, \mathbf{I_{\mathcal{T}_1}}, \dots, \mathbf{I_{\mathcal{S}_0}}, \mathbf{I_{\mathcal{S}_1}}, \dots\}.
\end{equation}

By revisiting target images from the past, we increase the diversity in the loss signal on the target domain and hence mitigate overfitting to the current scene. This further accounts for situations in which the current online image suffers from visual artifacts, e.g., overexposure. Similarly, revisiting samples from the source domain addresses the problem of catastrophic forgetting by ensuring that previously acquired knowledge can be preserved. Additionally, the annotations of the source samples enable pseudo-supervision on the target domain by exploiting cross-domain mixing strategies.
For both the target and the source replay, we randomly draw multiple samples from the respective replay buffer and apply augmentation to stabilize the loss. In particular, we perform RGB histogram matching of the source images to the online target image, and all available source samples have to be selected once before repetition is allowed to ensure diverse source supervision.

Similar to previous works~\cite{aljundi2019gradient, voedisch2023covio}, we explicitly consider limitations on the size of the replay buffer to closely resemble the deployment on a robotic platform, where disk storage is an important factor. This poses two questions: First, how to sample from $\mathcal{S}$ to construct the fixed source buffer $\mathbf{B_\mathcal{S}}$ that is prebuilt offline and, second, how to update the dynamic target buffer $\mathbf{B_\mathcal{T}}$ during deployment? To construct $\mathbf{B_\mathcal{S}}$, we propose a refined version of rare class sampling (RCS)~\cite{hoyer2022daformer}. The frequency $f_c$ of each class $c \in \mathcal{C}$ is calculated based on the number of pixels with class $c$:
\begin{equation}
    f_c = \frac{\sum_{\mathbf{I} \in \mathcal{S}} \sum_p^{H \times W} \mathbb{1}_c(p_{c'})}{|\mathcal{S}| \cdot H \cdot W},
\end{equation}
where $H$ and $W$ denote the height and width of the images in $\mathcal{S}$ and $p_{c'} \in \mathbf{I}$ refers to a pixel with class $c'$. The indicator function is 1 if $c'$ equals $c$ and 0 otherwise. The probability of sampling a class is then given by
\begin{equation}
    P(c) = \frac{e^{ (1 - f_c) / T }}{ \sum_{c' \in \mathcal{C}} e^{ (1 - f_{c'}) / T } },
\end{equation}
with temperature $T$ controlling the smoothness of the distribution, i.e., a smaller $T$ assigns a higher probability to rare classes. In detail, we first sample a class $c \sim P$ and then retrieve all candidate images containing pixels with class~$c$. Instead of taking a random image from these candidates, we sample according to the number of pixels with class~$c$. We repeat both steps $|\mathbf{B_\mathcal{S}}|$ times without selecting the same image more than once. Using RCS ensures that $\mathbf{B_\mathcal{S}}$ contains sufficiently many images with rare classes such that the performance on these classes will further improve during adaptation.

Since $\mathcal{T}$ does not contain annotations and, particularly in the beginning, predictions are not reliable, we cannot use RCS for updating $\mathbf{B_\mathcal{T}}$. Instead, we invert the common methodology of loop closure detection for visual SLAM~\cite{voedisch2023continual}, i.e., the image~$\mathbf{I_t}$ is only added to $\mathbf{B_\mathcal{T}}$ if its cosine similarity with respect to all samples within the buffer is below a threshold.
\begin{equation}
    \text{sim}_{\cos}(\mathbf{I_t}) = \max_{\mathbf{I_{\mathcal{T}_i}} \in \mathbf{B_\mathcal{T}}} \cos \left( \text{feat}(\mathbf{I_t}), \text{feat}(\mathbf{I_{\mathcal{T}_i}}) \right),
\end{equation}
where $\text{feat}(\cdot)$ refers to the image features extracted from the final layer of the shared encoder, which is not adapted. If $\mathbf{B_\mathcal{T}}$ is completely filled, we remove the following image to maximize image diversity:
\begin{equation}
    \argmax_{\mathbf{I_{\mathcal{T}_i}} \in \mathbf{B_\mathcal{T}}} \sum_{\mathbf{I_{\mathcal{T}_j}} \in \mathbf{B_\mathcal{T}}} \cos \left( \text{feat}(\mathbf{I_{\mathcal{T}_i}}), \text{feat}(\mathbf{I_{\mathcal{T}_j}}) \right)
\end{equation}
}


{\parskip=5pt
\noindent\textit{Depth Adaptation:}
To adapt the monocular depth estimation head along with the PoseNet, we exploit the fact that the photometric error loss (\refeqn{eqn:loss-photometric}) does not require ground truth annotations. Hence, we can directly transfer it to the implemented continual adaptation. In particular, we compute $\mathcal{L}^d_\mathit{pe}$ for the constructed batch~$\mathbf{b_t}$ and average the loss such that each sample contributes by the same amount:
\begin{equation}
    \mathcal{L}^d_\mathit{pe}(\mathbf{b_t}) = \frac{\mathcal{L}^d_\mathit{pe}(\mathbf{I_t}) + \sum_{i} \mathcal{L}^d_\mathit{pe}(\mathbf{I_{\mathcal{T}_i}}) + \sum_{j} \mathcal{L}^d_\mathit{pe}(\mathbf{I_{\mathcal{S}_j}}) }{ |\mathbf{b_t}| }.
\end{equation}

Furthermore, if the predicted camera motion is below a threshold, i.e., the robot is presumably not moving, we do not compute the $\mathcal{L}^d_\mathit{pe}(\mathbf{I_t})$ and subtract 1 from the denominator to avoid adding a bias to the current scene.
}


\begin{figure}[t]
    \centering
    \includegraphics[width=\linewidth]{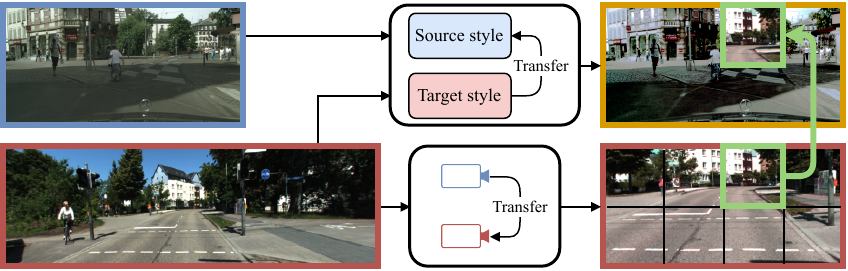}
    \vspace{-.6cm}
    \caption{Our proposed cross-domain mixing strategy first transfers the image style from the target to the source sample. Then it augments the target image to match the appearance of the source camera. Finally, a random image patch is copied from the target to the source image. The source annotations are retained and completed by the network's estimate on the copied image patch. The result serves as pseudo-label, combining self-iterative learning with ground truth supervision.}
    \label{fig:panoptic-mixup}
    \vspace{-.4cm}
\end{figure}

{\parskip=5pt
\noindent\textit{Panoptic Adaptation:}
As described in \refsec{ssec:ta-network-pretraining}, panoptic segmentation is the fused output of a semantic head and an instance head. We observe that the decrease in performance on samples from unseen domains can mostly be attributed to the semantic head, while instance predictions remain stable.
Cross-domain mixing strategies allow leveraging ideas from supervised training to an unsupervised setting, where ground truth annotations are unknown. In \net, we bootstrap annotated source samples and high-confident target predictions to artificially generate pseudo-labels for the target samples in an online fashion to supervise the semantic head. Similar to depth adaptation, we continue to compute $\mathcal{L}^\mathit{sem}_\mathit{bce}$ on $\{\mathbf{I_{\mathcal{S}_0}}, \mathbf{I_{\mathcal{S}_1}}, \dots\}$ to mitigate forgetting, and freeze the instance head.}

We further design a mixing strategy combining pixels of images from both $\mathcal{S}$ and $\mathcal{T}$, that considers multiple factors, which are unique to the online continual learning scenario: (1)~the robust pretraining on a dedicated source dataset, which may result in significant performance degradation on the target dataset if the pre-trained weights are strongly adapted; (2)~the existence of different cameras leading to significant changes in the field-of-view, geometric appearance of objects, resolution, and aspect ratio of the images; and (3)~the continuously evolving visual appearance of street scenes during adaptation.
To address these challenges, our cross-domain mixing approach employs a three-step method to generate the adaptation signal. First, we perform style transfer from the target image~$\mathbf{I_{\mathcal{T}_i}}$ to the source image~$\mathbf{I_{\mathcal{S}_j}}$ by aligning their pixel value histograms, as depicted in \reffig{fig:panoptic-mixup}. This allows supervision with ground truth labels on images that are of similar visual appearance as the target image. Second, we apply a geometric transformation on $\mathbf{I_{\mathcal{T}_i}}$ based on the camera intrinsics of the source and target domains denoted by $\mathbf{K_\mathcal{S}}$ and $\mathbf{K_\mathcal{T}}$, respectively. To this end, we assume a constant depth distribution over $\mathbf{I_{\mathcal{S}_j}}$, lift the pixel values into Euclidean space via inverse camera projection, and project the lifted points back into the camera view of $\mathbf{I_{\mathcal{T}_i}}$ as follows:
\begin{equation}
        \mathbf{I'_{\mathcal{T}}} (\mathbf{p_s}) = \mathbf{I_{\mathcal{T}}}
        \langle \mathbf{K_{\mathcal{T}}} \mathbf{K_{\mathcal{S}}^{-1}} \mathbf{p_s} \rangle, 
    \label{eqn:geom-augm}
\end{equation}
where $\langle\cdot\rangle$ denotes the bilinear sampling operator and $\mathbf{p_s}$ is a pixel coordinate in the source image.
\refEqn{eqn:geom-augm} results in an adapted target image $\mathbf{I'_\mathcal{T}}$ with an adjusted field of view, resolution, and a geometric appearance of the scene similar to that of $\mathbf{I_\mathcal{S}}$.
The final step in the process involves separating $\mathbf{I'_\mathcal{T}}$  into multiple segments and randomly selecting one of them to be inserted into the style-transferred source image, see \reffig{fig:panoptic-mixup}. To avoid providing a flawed supervision signal caused by geometrically unrealistic images, we only insert a single patch. 
Similarly, the ground truth labels of the pixels from $\mathbf{I_\mathcal{S}}$ are retained, and the semantic labels estimated by the network are used to label the inserted patch after intrinsics transformation.
The generated image is then fed into the network and training is performed using the cross-entropy loss and the generated pseudo-labels of the mixed image.
To mitigate the decline in performance commonly associated with self-iterative training on predicted pseudo-labels, often resulting in class collapse, we utilize an exponentially moving average~(EMA) filter for updating the network weights.
In detail, we create a duplicate with network weights $w_\text{EMA}$ of the initial model with weights $w$ and use this so-called EMA model to generate the semantic predictions.
During continual learning, the weights $w$ are updated via backpropagation on $\mathbf{b_t}$. Then, the EMA model is updated as follows:
\begin{equation}
    w_\text{EMA} \leftarrow \alpha \cdot w_\text{EMA} + (1 - \alpha) \cdot w,
    \label{eqn:ema}
\end{equation}
where $\alpha$ denotes the contribution of the EMA model.

\section{Experimental Evaluation}

In the following sections, we provide further details on the pretraining step and the datasets that we evaluate on. We present extensive experimental results on the efficiency and efficacy of our proposed approach and include ablation studies on important design choices. Finally, we expand the experimental setup to multi-domain adaptation closely resembling classical continual learning settings.

We follow the evaluation protocol of \citet{zhang2020online}. In detail, we compute the evaluation metrics on the frame of the current timestamp before using the same frame to perform backpropagation and update the model weights. Once \SI{70}{\percent} of a sequence is processed, we calculate the average of the accumulated metrics. Additionally, we report the scores on the remaining \SI{30}{\percent} of the same sequence without further weight updates to analyze the performance of the adapted model. In the tables, we refer to these types of evaluation by protocol~1 and protocol~2, respectively. We further denote the respective parts of a sequence by \textit{adapt} and \textit{eval}.
Unlike \citet{zhang2020online}, we define our task in the context of continual learning. To measure knowledge retention and hence mitigate catastrophic forgetting, we introduce protocol~3 as evaluating the adapted model on the \textit{val} split of the source dataset.


\subsection{Datasets}
To simulate data from a variety of domains, we employ our method on three datasets, namely Cityscapes~\cite{cordts2016the}, KITTI-360~\cite{liao2022kitti360}, and SemKITTI-DVPS~\cite{behley2019semantickitti}. In particular, we utilize Cityscapes for pre-training and sequences of both KITTI-360 and SemKITTI-DVPS for adaptation. In the supplementary video we further provide qualitative results on in-house data recorded with our robotic platform.


{\parskip=3pt
\noindent\textit{Cityscapes:}
The Cityscapes Dataset~\cite{cordts2016the} is a large-scale autonomous driving dataset that was recorded in 50 cities in Germany and bordering regions. It includes RGB images, panoptic annotations, and vehicle metadata. In this work, we utilize the fine panoptic labels to train the semantic and instance heads in a supervised manner. Additionally, we leverage the sequence image data of the left camera to train the depth prediction in an unsupervised fashion. Finally, we compute the depth error metrics using the provided disparity maps.
}


{\parskip=3pt
\noindent\textit{KITTI-360:}
The KITTI-360 Dataset~\cite{liao2022kitti360} is a relatively recently released public dataset for the domain of autonomous driving, which was recorded in the city of Karlsruhe, Germany. It includes both 2D and 3D panoptic annotations for RGB images and LiDAR data. In this work, we predominantly utilize the RGB images to simulate an online image stream of an onboard camera. In particular, we use these images to adapt our network in a self-supervised manner. To compute evaluation metrics, we compare our predictions with the ground truth measurements and annotations of the dataset.
}


{\parskip=3pt
\noindent\textit{SemKITTI-DVPS:}
The SemKITTI-DVPS~\cite{qiao2021vipdeeplab} is based on the odometry benchmark of the KITTI Dataset~\cite{geiger2012are}, which was recorded in Karlsruhe, Germany. We utilize the RGB images to simulate an onboard camera and to adapt our network to the new domain. Furthermore, we compute depth metrics based on the provided projected LiDAR points and the semantic/panoptic metrics using the extension SemanticKITTI~\cite{behley2019semantickitti}.
}


{\parskip=3pt
\noindent\textit{Semantic Labels:}
As the aforementioned datasets use different labeling policies for the semantic annotations, we use the 19 classes of Cityscapes as the reference definition and remap classes of the other datasets. However, certain classes do not exist in the adaptation datasets (\textit{wall}, \textit{traffic light}, \textit{bus}, \textit{train}). For consistency across the datasets, we merge \textit{wall} with \textit{building} and remove the other three classes. Additionally, we merge \textit{motorcycle} and \textit{bicycle} into \textit{two-wheeler} to increase the number of annotated pixels. Consequently, we consider nine ``stuff'' classes and five ``thing'' classes, listed in \reftab{tab:classwise-evaluation}. Note that \textit{sky} is not included in SemKITTI-DVPS due to using LiDAR annotations and hence excluded in the evaluation on this dataset.
}


\begin{table}
\scriptsize
\centering
\caption{Efficacy of the Network}
\vspace{-0.2cm}
\label{tab:comparison-with-guda}
\setlength\tabcolsep{4.8pt}
\begin{threeparttable}
    \begin{tabular}{l | c | c ccc}
        \toprule
        \textbf{Method} & \textbf{Dataset} & \textbf{mIoU} $\uparrow$ & \textbf{RMSE} $\downarrow$ & \textbf{Abs Rel} $\downarrow$ & $\mathbf{\delta_1}$ $\uparrow$ \\
        \midrule
        GUDA & \multirow{2}{*}{KITTI} & 
                        ---  & 4.42 & 0.11 & 0.88 \\
        \net & & 62.8 & 3.52 & 0.09 & 0.90 \\
        \midrule
        GUDA & \multirow{2}{*}{Cityscapes} & 
                        72.9  & ---   & ---  & ---  \\
        \net & & 72.9 & 10.16 & 0.19 & 0.78 \\
        \bottomrule
    \end{tabular}
    \footnotesize
    Our utilized network is able to reproduce the performance of the baseline method GUDA~\cite{guizilini2021geometric} for both semantic segmentation (mIoU) and depth estimation (RMSE, Abs Rel, $\delta_1$).
    The performance of GUDA is reported by the authors. To evaluate \net on KITTI, we use sequence 08 \textit{eval} of SemKITTI-DVPS.
\end{threeparttable}
\vspace{-0.5cm}
\end{table}

\subsection{Pretraining Protocol}

The initial state of the network weights before adaptation is obtained by initializing the encoders using pretrained weights from the ImageNet dataset, followed by training the entire model on the Cityscapes dataset. In detail, we use the Adam optimizer with a constant learning rate $lr = 0.0001$ and train the entire network for 250 epochs. In our experiments, we compare the performance of our approach to directly training on the target dataset, which can be considered as a theoretical upper limit having full target knowledge. Due to the unbalanced class distribution of KITTI-360, we train a copy of the network in two steps, using the Adam optimizer with $lr = 0.0001$ on sequences \mbox{$00$-$07$}. We train for 45 epochs while ignoring the most common classes \textit{road}, \textit{sidewalk}, \textit{building}, and \textit{vegetation}, followed by 55 epochs including all classes. Similarly, for SemKITTI-DVPS, we train another copy of the network on sequences $00$-$06$, $09$, and $10$ for 30 epochs without the aforementioned classes plus \textit{terrain} and \textit{sky}, which is not included in the dataset, followed by 30 epochs including all classes.
In \reftab{tab:comparison-with-guda}, we demonstrate that our implemented network is able to reproduce the performance of the baseline method GUDA~\cite{guizilini2021geometric}.


\begin{table*}[t]
\scriptsize
\centering
\caption{Adaptation Performance}
\vspace{-0.2cm}
\label{tab:baselines}
\begin{threeparttable}
    \begin{tabular}{l | c | c ccc cc | c ccc cc}
        \toprule
        \multirow{2}{*}{\textbf{Method}} & \multirow{2}{*}{\textbf{Sequence}} & \multicolumn{6}{c|}{\textbf{Protocol 1}} & \multicolumn{6}{c}{\textbf{Protocol 2}} \\
        & & mIoU $\uparrow$ & PQ $\uparrow$ & SQ $\uparrow$ & RQ $\uparrow$ & RMSE $\downarrow$ & Abs Rel $\downarrow$ & mIoU $\uparrow$ & PQ $\uparrow$ & SQ $\uparrow$ & RQ $\uparrow$ & RMSE $\downarrow$ & Abs Rel $\downarrow$ \\
        \midrule
                Only source & \multirow{2}{*}{00} & 51.61 & 39.10 & 72.72 & 50.48 & 6.54 & 0.36 & 49.94 & 35.29 & 72.14 & 45.50 & 6.08 & 0.34 \\
        \net      & & \textbf{53.76} & \textbf{40.72} & \textbf{72.90} & \textbf{52.51} & \textbf{5.09} & \textbf{0.19} & \textbf{52.08} & \textbf{36.08} & \textbf{72.58} & \textbf{46.08} & \textbf{4.34} & \textbf{0.15} \\
        \midrule
        Only source & \multirow{2}{*}{02} & 45.97 & 31.83 & 67.62 & 41.08 & 6.26 & 0.35 & 46.55 & 30.13 & 65.03 & 39.30 & 6.06 & 0.36 \\
        \net      & & \textbf{46.62} & \textbf{32.11} & \textbf{67.74} & \textbf{41.62} & \textbf{4.31} & \textbf{0.16} & \textbf{47.48} & \textbf{30.33} & \textbf{65.35} & \textbf{39.46} & \textbf{3.76} & \textbf{0.13} \\
        \midrule
        Only source & \multirow{2}{*}{03} & 46.63 & 28.15 & 57.41 & 35.23 & \textbf{8.20} & 0.34 & \textbf{52.10} & 28.20 & 56.67 & 35.77 & 7.34 & 0.29 \\
        \net      & & \textbf{47.94} & \textbf{29.05} & \textbf{58.07} & \textbf{36.10} & 8.26 & \textbf{0.33} & 52.00 & \textbf{31.13} & \textbf{61.51} & \textbf{39.65} & \textbf{6.98} & \textbf{0.18} \\
        \midrule
        Only source & \multirow{2}{*}{04} & 45.02 & 29.34 & 65.48 & 38.15 & 6.70 & 0.37 & 45.53 & 30.13 & \textbf{70.85} & 38.89 & 6.61 & 0.38 \\
        \net      & & \textbf{45.40} & \textbf{29.78} & \textbf{65.89} & \textbf{38.84} & \textbf{5.00} & \textbf{0.19} & \textbf{45.68} & \textbf{30.63} & 66.18 & \textbf{39.89} & \textbf{4.33} & \textbf{0.17} \\ 
        \midrule
        Only source & \multirow{2}{*}{05} & 48.94 & 32.19 & 66.80 & 41.37 & 6.76 & 0.37 & \textbf{44.52} & \textbf{27.34} & \textbf{60.72} & \textbf{35.58} & 5.93 & 0.43 \\
        \net      & & \textbf{49.26} & \textbf{32.96} & \textbf{66.98} & \textbf{42.40} & \textbf{5.25} & \textbf{0.21} & 43.79 & 26.48 & 60.33 & 34.88 & \textbf{4.68} & \textbf{0.25} \\ 
        \midrule
        Only source & \multirow{2}{*}{06} & 46.03 & 29.88 & 66.58 & 38.42 & 6.09 & 0.39 & 46.28 & 31.79 & 70.47 & 41.40 & 6.12 & 0.37 \\
        \net      & & \textbf{46.53} & \textbf{30.45} & \textbf{66.66} & \textbf{39.20} & \textbf{4.97} & \textbf{0.22} & \textbf{47.27} & \textbf{31.99} & \textbf{70.74} & \textbf{41.71} & \textbf{4.23} & \textbf{0.18} \\ 
        \midrule
        Only source & \multirow{2}{*}{07} & 40.54 & 28.48 & 66.52 & 34.42 & 7.83 & 0.34 & 59.07 & 27.62 & 45.88 & 35.41 & 9.64 & 0.38 \\
        \net      & & \textbf{41.46} & \textbf{29.30} & \textbf{67.64} & \textbf{35.58} & \textbf{6.50} & \textbf{0.22} & \textbf{60.57} & \textbf{30.91} & \textbf{50.25} & \textbf{39.79} & \textbf{6.48} & \textbf{0.20} \\
        \midrule
        Only source & \multirow{2}{*}{09} & 50.59 & 37.26 & 74.06 & 47.38 & 6.03 & 0.36 & 50.78 & 36.57 & 72.22 & 46.75 & 5.60 & 0.35 \\
        \net      & & \textbf{52.29} & \textbf{38.02} & \textbf{74.88} & \textbf{48.21} & \textbf{4.74} & \textbf{0.19} & \textbf{51.53} & \textbf{37.56} & \textbf{72.87} & \textbf{47.99} & \textbf{4.56} & \textbf{0.16} \\
        \midrule
        Only source & \multirow{2}{*}{10} & 51.94 & 32.60 & 71.27 & 32.60 & 8.06 & 0.35 & 45.74 & 30.62 & 69.56 & 39.49 & 7.90 & 0.33 \\
        \net      & & \textbf{53.02} & \textbf{33.50} & \textbf{71.53} & \textbf{33.50} & \textbf{7.19} & \textbf{0.22} & \textbf{49.91} & \textbf{31.91} & \textbf{70.68} & \textbf{40.95} & \textbf{5.57} & \textbf{0.15} \\
        \bottomrule
    \end{tabular}
    \footnotesize
    Comparison between our \net and the performance of the same architecture without performing online continual learning on the respective sequence of the KITTI-360 dataset. Thus, ``only source'' refers to the model weights after pretraining on Cityscapes. The listed metrics are mean intersection over union (mIoU) for semantic segmentation; panoptic quality (PQ), segmentation quality (SQ), and recognition quality (RQ) for panoptic segmentation; root mean squared error (RMSE) and absolute relative error (Abs Rel) for monocular depth estimation. Bold values denote the best result on each sequence.
\end{threeparttable}
\vspace*{-.2cm}
\end{table*}

\begin{table*}[t]
\scriptsize
\centering
\caption{Continual Learning for Monocular Depth Estimation}
\vspace{-0.2cm}
\label{tab:depth-adaptation}
\setlength\tabcolsep{3.5pt}
\begin{threeparttable}
    \begin{tabular}{l | c | cc ccc | cc ccc | cc ccc}
        \toprule
        \multirow{2}{*}{\textbf{Method}} & \textbf{Batch} & \multicolumn{5}{c|}{\textbf{Protocol 1}} & \multicolumn{5}{c|}{\textbf{Protocol 2}} & \multicolumn{5}{c}{\textbf{Protocol 3}} \\
        & current/target/source & RMSE & Abs Rel & $\delta_1$ & $\delta_2$ & $\delta_3$ & RMSE & Abs Rel & $\delta_1$ & $\delta_2$ & $\delta_3$ & RMSE & Abs Rel & $\delta_1$ & $\delta_2$ & $\delta_3$ \\
        \midrule
        Only target & 0 / 0 / 0 & 6.13 & 0.15 & 0.84 & 0.93 & 0.96 & 4.78 & 0.12 & 0.88 & 0.95 & 0.97 & 12.22 & 0.26 & 0.51 & 0.82 & 0.94 \\
        Only source & 0 / 0 / 0 & 8.06 & 0.35 & 0.43 & 0.77 & 0.91 & 7.90 & 0.33 & 0.44 & 0.77 & 0.93 & \textbf{10.16} & \textbf{0.19} & \textbf{0.78} & \textbf{0.93} & \textbf{0.97} \\
        \midrule
        Online image & 1 / 0 / 0 & 8.33 & 0.27 & 0.64 & 0.84 & 0.93 & 6.06 & 0.33 & 0.46 & 0.73 & 0.90 & 13.72 & 0.57 & 0.30 & 0.50 & 0.68 \\
        Target replay & 1 / 2 / 0 & \textbf{6.35} & \textbf{0.19} & \textbf{0.77} & \textbf{0.91} & \textbf{0.96} & \textbf{5.34} & \textbf{0.15} & \textbf{0.81} & \textbf{0.93} & \textbf{0.97} & 12.48 & 0.44 & 0.34 & 0.68 & 0.88 \\ 
        \rowcolor{Gray}
        \net & 1 / 2 / 2 & \underline{7.19} & \underline{0.22} & \underline{0.73} & \underline{0.89} & \underline{0.94} & \underline{5.57} & \textbf{0.15} & \textbf{0.81} & \textbf{0.93} & \textbf{0.97} & \underline{11.38} & \underline{0.21} & \underline{0.75} & \underline{0.91} & \underline{0.96} \\ 
        \bottomrule
    \end{tabular}
    \footnotesize
    The root mean squared error (RMSE), absolute relative error (Abs Rel) as well as accuracies $\delta_1 = \delta < 1.25$, $\delta_2 = \delta < 1.25^2$, and $\delta_3 = \delta < 1.25^3$, obtained by adapting Cityscapes to sequence 10 of the KITTI-360 dataset. Best results without access to ground truth target data (``only target'') in each category are in \textbf{bold}; second best are \underline{underlined}.
\end{threeparttable}
\vspace*{-.3cm}
\end{table*}

\begin{table*}
\scriptsize
\centering
\caption{Continual Learning for Panoptic Segmentation}
\vspace{-0.2cm}
\label{tab:panoptic-adaptation}
\setlength\tabcolsep{3.7pt}
\begin{threeparttable}
    \begin{tabular}{l | c ccc | c ccc | c ccc}
        \toprule
        \multirow{2}{*}{\textbf{Method}} & \multicolumn{4}{c|}{\textbf{Protocol 1}} & \multicolumn{4}{c|}{\textbf{Protocol 2}} & \multicolumn{4}{c}{\textbf{Protocol 3}} \\
        & mIoU & PQ & SQ & RQ & mIoU & PQ & SQ & RQ & mIoU & PQ & SQ & RQ \\
        \midrule
        Only target & 64.65 & 41.91 & 76.68 & 51.48 & 55.12 & 36.58 & 67.41 & 46.15 & 46.33 & 28.24 & 70.03 & 36.92 \\
        Only source & 51.94 & 32.60 & 71.27 & 42.44 & 45.74 & 30.62 & 69.56 & 39.49 & \underline{72.87} & 49.19 & \underline{77.45} & 60.40 \\
        \midrule
        GUDA~\cite{guizilini2021geometric} & 45.56 & 29.70 & 70.67 & 39.05 & 47.62 & 31.03 & 64.00 & 40.49 & 66.57 & 44.39 & 75.95 & 55.32 \\        
        DACS~\cite{tranheden2021dacs} & 51.14 & 32.09 & 71.12 & 42.23 & 45.24 & 29.05 & 69.47 & 38.11 & 72.66 & 49.27 & 77.33 & 60.60\\        
        \midrule
        \net (online image) & \textbf{53.22} & \underline{33.46} & \textbf{71.63} & \underline{43.46} & \underline{49.51} & 31.49 & 64.17 & \underline{40.71} & 72.81 & \textbf{49.83} & 77.25 & \textbf{61.49} \\
        \net (random sampling) & 52.36 & 33.24 & \underline{71.60} & 43.25 & 48.78 & \underline{31.50} & \underline{68.83} & 40.56 & 72.05 & 49.11 & 77.18 & 60.52 \\
        \midrule
        \rowcolor{Gray}
        \net & \underline{53.02} & \textbf{33.50} & 71.53 & \textbf{43.62} & \textbf{49.91} & \textbf{31.91} & \textbf{70.68} & \textbf{40.95} & \textbf{72.90} & \underline{49.76} & \textbf{77.49} & \underline{61.22} \\
        \bottomrule
    \end{tabular}
    \footnotesize
    The mean intersection over union (mIoU), panoptic quality (PQ), semantic quality (SQ), and recognition quality (RQ) are obtained by adapting Cityscapes to sequence 10 of the KITTI-360 dataset. Best results without access to ground truth target data (``only target'') in each category are in \textbf{bold}; second best are \underline{underlined}.
\end{threeparttable}
\vspace*{-.3cm}
\end{table*}

\subsection{Online Adaptation}
\label{ssec:exp-online-adaptation}

In this section, we extensively evaluate our proposed \net with respect to both adapting to a new domain and retaining knowledge to mitigate forgetting. In detail, for all presented experiments, we freeze the shared encoder following the study by \citet{mccraith2020monocular}. Based on the ablation study in \refsec{ssec:exp-ablation-study-replay-buffer}, we use a buffer size of 300. For RCS, we follow \citet{hoyer2022daformer} and set $T=0.01$. Updating the EMA model is done with $\alpha = 0.99$.

In \reftab{tab:baselines}, we assess the performance of \net on all sequences of the KITTI-360 dataset and compare it with the baseline method ``only source'', which is also pretrained on Cityscapes but does not perform further adaptation to the target domain $\mathcal{T}$. This approach should be interpreted as a lower performance bound that must be improved.
We demonstrate the key performance metrics of both protocols~1 and 2.
As shown in \reftab{tab:baselines}, \net achieves a performance boost across the board, as measured by the mIoU metric and all depth metrics of protocol~1. We attribute this improvement to the additional supervision signals incorporated into the segmentation head through our mixing strategy and the self-supervised reconstruction loss for depth adaptation. The improvement in semantic segmentation further enhances the panoptic segmentation metrics.
With respect to protocol~2, \net reduces the depth errors on all sequences and improves the performance of semantic and panoptic segmentation on the vast majority of sequences. Note that on sequence~03 the panoptic metrics increase significantly despite the consistent mIoU, which we attribute to the more refined segmentation of objects due to our proposed cross-domain mixing strategy.

\begin{figure*}
    \centering
    \includegraphics[width=\linewidth]{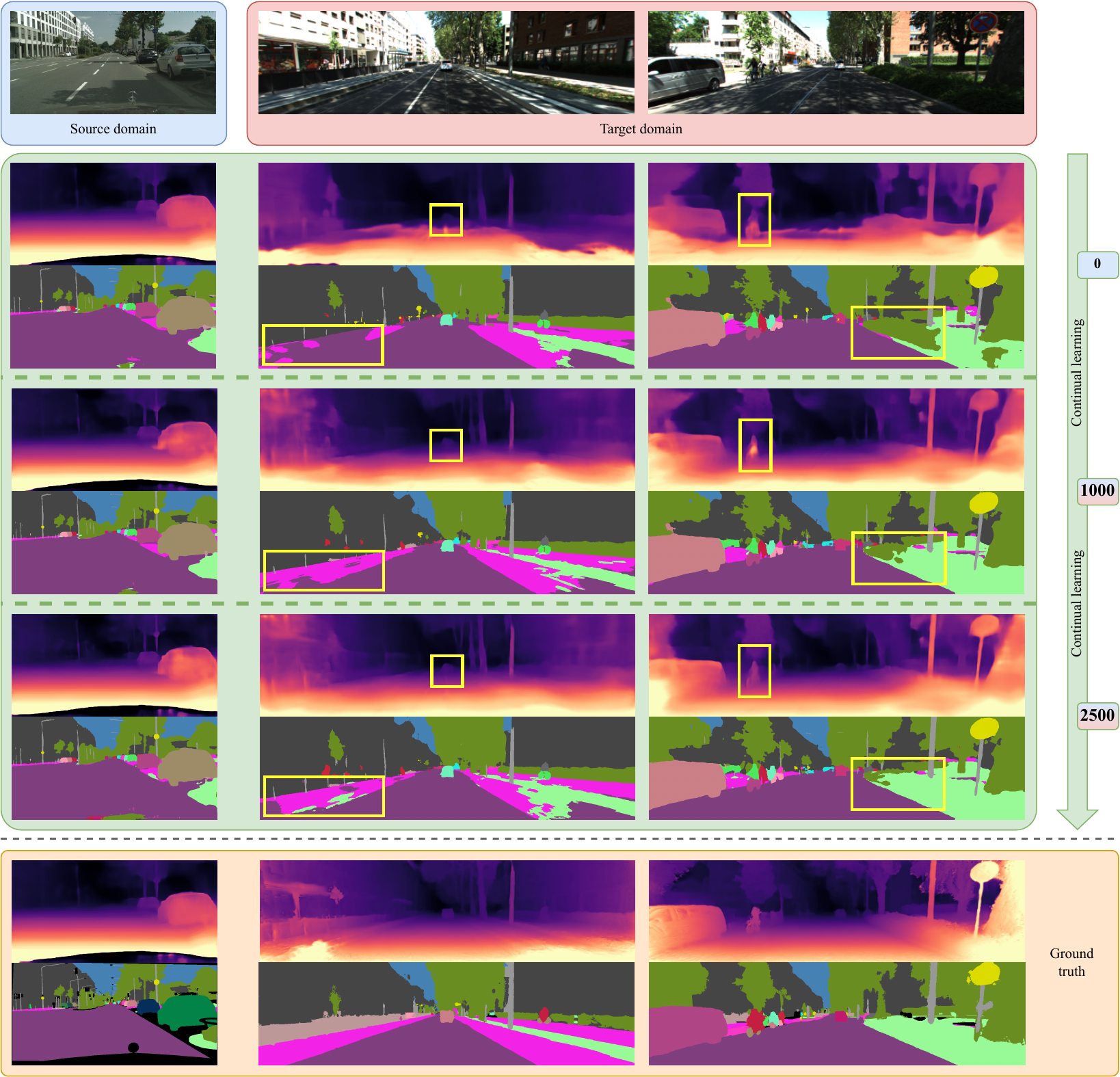}
    \vspace*{-.5cm}
    \caption{Qualitative results for Cityscapes to KITTI-360 adaptation after pretraining on the source, i.e., 0 steps, and after having seen 1,000 and 2,500 frames. As shown in the left column, \net is able to avoid catastrophic forgetting on the source domain. The progressive adaptation on the target domain is particularly visible in the image areas highlighted by yellow boxes. ``Stuff'' classes of similar appearance like \textit{sidewalk} vs. \textit{road} (left image) and \textit{terrain} vs. \textit{vegetation} (right image) can be better distinguished by \net. Furthermore, instances become more pronounced as can be observed for the highlighted car (left image) and the cyclist (right image).
    }
    \label{fig:qualitative-results}
\end{figure*}

For the following experiments, we consider the case of using Cityscapes as the source domain and sequence~10 of \mbox{KITTI-360} as the target domain. In \reffig{fig:qualitative-results}, we illustrate the adaptation progress using unseen validation samples and compare the results to the ground truth. For depth, we visualize predictions generated by the network if it was only trained on $\mathcal{S}$ and $\mathcal{T}$, respectively.
For panoptic segmentation, the progressive adaptation on the target domain is particularly visible on the \textit{sidewalk} and \textit{terrain} image regions, which \net learns to differentiate from the similarly looking classes \textit{road} and \textit{vegetation}. Furthermore, instances become more pronounced, e.g., the cyclist in the right sample. Despite the enhancements on the target domain, \net successfully maintains its performance on the source domain with only minimum decreases in depth estimation.


{\parskip=5pt
\noindent\textit{Depth Adaptation:}
We present the results for monocular depth estimation in \reftab{tab:depth-adaptation}. The first row ``only target'' shows the theoretical performance on $\mathcal{T}$ (protocols~1 and 2) if the network would have been trained directly on this domain. Note that such a setup is infeasible in the real world when continuous operation must be guaranteed. The second row ``only source'' denotes the performance after pretraining on~$\mathcal{S}$ without performing online continual learning. Comparing the absolute relative error as well as the accuracies $\delta_1$, $\delta_2$, and~$\delta_3$ between these rows reveals the domain gap. Note that the opposite gap can be observed when evaluating on~$\mathcal{S}$ (protocol~3). While continual learning using the current online sample increases the accuracy of protocol~1, it also overfits to the current scene. That is, generalizability to the entire target domain is not achieved as shown by protocol~2. Introducing replay samples from the target buffer overcomes this issue and accounts for online samples of poor quality, improving protocols~1 and 2. However, both of the above result in catastrophic forgetting with respect to $\mathcal{S}$ (protocol~3). The final \net adds additional source replay yielding low errors and high accuracy by compromising on both $\mathcal{S}$ and $\mathcal{T}$.
}


{\parskip=5pt
\noindent\textit{Panoptic Adaptation:}
In \reftab{tab:panoptic-adaptation}, we also demonstrate the domain gap between $\mathcal{S}$ and $\mathcal{T}$ for semantic and panoptic segmentation. Similar to depth estimation, both ``only target'' and ``only source'' only perform well on their respective training domain without being able to generalize to the other.
We further evaluate \net by comparing it with two competitive baselines that perform domain adaptation on segmentation tasks: GUDA~\cite{guizilini2021geometric}, which combines semantic segmentation and depth estimation, rendering their task comparable to ours, and DACS~\cite{tranheden2021dacs}, which employs a class-mix strategy for offline domain adaptation of semantic segmentation. To ensure a fair comparison, both baselines are evaluated using the same settings as \net, including diversity sampling-based experience replay. The results in \reftab{tab:panoptic-adaptation} indicate that both approaches lead to a significant performance decrease across all three protocols. GUDA's reliance on self-supervised feature alignment using depth training is not effective in the continual learning setting, as shown in the results. DACS also suffers from a decline in performance, likely due to the strong intervention of its mixing strategy into the pretrained network, which can already produce reasonable predictions on the target domain without adaptation. 

These results imply that traditional approaches from offline sim-to-real adaptation may not perform well in the online continual learning scenario.
To further assess the impact of target replay and our diversity-based buffer sampling, we selectively deactivate both components. Applying the proposed cross-domain mixing strategy results in an improvement in protocol~1. However, similar to depth adaptation, the results are not fully generalizable to the entire target domain, e.g., SQ of protocol~2. Instead of diversity-based sampling, we use random sampling when both creating the source buffer and when updating the target buffer. Compared to \net, the performance heavily degrades demonstrating the efficacy of the sampling method.
Finally, we present the classwise evaluation of the segmentation performance in \reftab{tab:classwise-evaluation}, which demonstrates improvements of \net in the IoU metrics for most classes. In particular, we observe significant enhancements of the \textit{two-wheeler} and \textit{terrain} classes. The latter can also be observed in \reffig{fig:qualitative-results}. In fact, \net outperforms even the model trained directly on the target domain using ground truth supervision for the latter class.
}

\begin{table}
\scriptsize
\centering
\caption{Classwise evaluation}
\vspace{-0.2cm}
\label{tab:classwise-evaluation}
\begin{threeparttable}
    \begin{tabular}{c | l | c c c}
        \toprule
        \multicolumn{2}{c|}{\textbf{Class}} & \textbf{Only target} & \textbf{Only source} & \textbf{\net} \\
        \midrule
        \multirow{9}{*}{\begin{sideways} Stuff \end{sideways}}
        & Road          & 93 & 89 & 91 \\
        & Sidewalk      & 40 & 32 & 37 \\
        & Building      & 88 & 85 & 85 \\
        & Fence         & 43 & 14 & 22 \\
        & Pole          & 35 & 29 & 32 \\
        & Traffic sign  & 40 & 35 & 38 \\
        & Vegetation    & 78 & 73 & 75 \\
        & Terrain       & 54 & 21 & 39 \\
        & Sky           & 82 & 79 & 81 \\
        \midrule
        \multirow{5}{*}{\begin{sideways} Thing \end{sideways}}
        & Person        & 47 & 38 & 38 \\
        & Rider         & 47 & 29 & 36 \\
        & Car           & 91 & 83 & 84 \\
        & Truck         &  1 &  4 &  2 \\
        & Two-wheeler   & 33 & 27 & 38 \\
        \midrule
        \multicolumn{2}{c|}{Mean} & 55.1 & 45.7 & 49.9 \\
        \bottomrule
    \end{tabular}
    \footnotesize
    The classwise mIoU is based on protocol 2 in \reftab{tab:panoptic-adaptation}. We compare \net against two baselines that were trained using source (``only source'') or target data (``only target''), respectively. \net provides a significant performance boost of 4.2\% in terms of the mIoU metric.
\end{threeparttable}
\vspace*{-.3cm}
\end{table}


\subsection{Ablation Study of the Replay Buffer}
\label{ssec:exp-ablation-study-replay-buffer}

We extensively study different sizes of the replay buffer and the effect of diversity sampling as explained in \refsec{ssec:ta-online-adaptation}. We list our results in \reftab{tab:ablation-buffer-size}. Note that an infinite replay buffer contains 2,975 source and a maximum of 2,683 target samples in the employed setting, i.e., adapting from Cityscapes \textit{train} to KITTI-360 using sequence 10 \textit{adapt}. Generally, a larger replay buffer yields higher performance with respect to both adaptation capability and avoiding catastrophic forgetting. Additionally, the proposed diversity sampling using semantic classes for the source and image features for the target samples increases the performance throughout the experiments. However, a greater buffer size increases the required storage posing a challenge for real-world deployment. Based on the presented results, we select a buffer size of 300 with active diversity sampling as for smaller sizes the performance of semantic segmentation on the target domain degrades.

\begin{table*}
\scriptsize
\centering
\caption{Ablation Study on the Replay Buffer}
\vspace{-0.2cm}
\label{tab:ablation-buffer-size}
\setlength\tabcolsep{2.7pt}
\begin{threeparttable}
    \begin{tabular}{c | c | c ccc cc | c ccc cc}
        \toprule
        \multirowcell{2}{\textbf{Size}} & \multirowcell{2}{\textbf{Div.}} & \multicolumn{6}{c|}{\textbf{Protocol 2}} & \multicolumn{6}{c}{\textbf{Protocol 3}}  \\
        & & mIoU $\uparrow$ & PQ $\uparrow$ & SQ $\uparrow$ & RQ $\uparrow$ & RMSE $\downarrow$ & Abs Rel $\downarrow$ & mIoU $\uparrow$ & PQ $\uparrow$ & SQ $\uparrow$ & RQ $\uparrow$ & RMSE $\downarrow$ & Abs Rel $\downarrow$ \\
        \midrule
        $\infty$ &    & 49.15 & 31.95 & 69.08 & 40.96 & 4.94 & 0.15 & 73.25 & 50.37 & 77.77 & 61.87 & 10.76 & 0.21 \\
        \midrule
        1000 &        & 49.11$\pm$0.69 & \underline{31.85}$\pm$0.25 & 66.82$\pm$3.06 & \underline{40.93}$\pm$0.06 & \textbf{5.04}$\pm$0.01 & \textbf{0.14}$\pm$0.00 & 72.84$\pm$0.33 & \underline{49.93}$\pm$0.20 & \underline{77.51}$\pm$0.05 & \underline{61.39}$\pm$0.28 & 11.35$\pm$0.39 & \underline{0.22}$\pm$0.01 \\
        1000 & \cmark & 49.36 & 31.83 & 68.89 & \textbf{41.01} & 5.30 & \underline{0.15} & \textbf{73.50} & \textbf{50.05} & \textbf{77.67} & \textbf{61.48} & 12.06 & 0.23 \\ 
        500 &         & 48.77$\pm$0.39 & 31.54$\pm$0.39 & 67.39$\pm$2.16 & 40.66$\pm$0.54 & \underline{5.20}$\pm$0.20 & \underline{0.15}$\pm$0.00 & 72.38$\pm$0.26 & 49.48$\pm$0.14 & 77.45$\pm$0.16 & 60.90$\pm$0.23 & \textbf{11.14}$\pm$0.54 & \underline{0.22}$\pm$0.01 \\
        500 & \cmark  & \underline{49.56} & 31.83 & 70.11 & 40.96 & 5.55 & 0.16 & 72.78 & 49.68 & 77.39 & 61.10 & \underline{11.30} & \underline{0.22} \\
        300 &         & 48.78$\pm$0.05 & 31.50$\pm$0.19 & \underline{68.83}$\pm$2.31 & 40.56$\pm$0.15 & 5.27$\pm$0.16 & \underline{0.15}$\pm$0.00 & 72.05$\pm$0.25 & 49.11$\pm$0.30 & 77.18$\pm$0.07 & 60.52$\pm$0.35 & \textbf{11.14}$\pm$0.22 & \underline{0.22}$\pm$0.01 \\
        \rowcolor{Gray}
        300 & \cmark  & \textbf{49.91} & \textbf{31.91} & \textbf{70.68} & 40.95 & 5.57 & \underline{0.15} & \underline{72.90} & 49.76 & 77.49 & 61.22 & 11.38 & \textbf{0.21} \\
        100 &         & 48.27$\pm$0.84 & 30.71$\pm$0.41 & 63.95$\pm$0.41 & 39.79$\pm$0.38 & 5.83$\pm$0.15 & 0.16$\pm$0.00 & 69.75$\pm$1.77 & 47.94$\pm$0.95 & 76.66$\pm$0.25 & 59.39$\pm$1.16 & 10.86$\pm$0.56 & \underline{0.22}$\pm$0.02 \\
        100 & \cmark  & 48.40 & 30.85 & 64.07 & 39.95 & 5.31 & 0.16 & 72.35 & 48.81 & 77.16 & 60.25 & 11.71 & \underline{0.22} \\
        25 &          & 46.03$\pm$1.03 & 29.62$\pm$0.37 & 66.10$\pm$2.26 & 38.48$\pm$0.45 & 5.25$\pm$0.26 & \textbf{0.14}$\pm$0.01 & 67.23$\pm$0.85 & 45.90$\pm$0.66 & 75.69$\pm$0.38 & 57.21$\pm$0.76 & 11.81$\pm$0.22 & \underline{0.22}$\pm$0.01 \\
        25 & \cmark   & 46.35 & 29.73 & 63.35 & 38.58 & 5.62 & 0.17 & 68.84 & 46.34 & 76.06 & 57.78 & 12.51 & 0.24 \\
        \bottomrule
    \end{tabular}
    \footnotesize
    The numbers above are obtained by adapting Cityscapes to sequence 10 of the KITTI-360 dataset. Here, an infinite buffer size equals 2,975 source samples and a maximum of 2,683 target samples.
    Note that the effective size is two times the shown value as it refers to both source and target replay. The term ``Div.'' refers to diversity sampling. Where diversity sampling is not used, the same experiment is repeated three times with different random seeds to ensure a statistically reliable measure of performance. The results of these experiments are presented as the mean and standard deviation. Best results in each category are in \textbf{bold}; second best are \underline{underlined}.
\end{threeparttable}
\vspace*{-.3cm}
\end{table*}


\begin{table*}
\scriptsize
\centering
\caption{Continual Learning on Multiple Domains}
\vspace{-0.2cm}
\label{tab:continual-learning}
\setlength\tabcolsep{2.7pt}
\begin{threeparttable}
    \begin{tabular}{l | c ccc cc | c ccc cc | c ccc cc}
        \toprule
        \textbf{Domain} & \textbf{mIoU} & \textbf{PQ} & \textbf{SQ} & \textbf{RQ} & \textbf{RMSE} & \textbf{Abs Rel} & \textbf{mIoU} & \textbf{PQ} & \textbf{SQ} & \textbf{RQ} & \textbf{RMSE} & \textbf{Abs Rel} & \textbf{mIoU} & \textbf{PQ} & \textbf{SQ} & \textbf{RQ} & \textbf{RMSE} & \textbf{Abs Rel} \\
        \midrule
        & \multicolumn{5}{r}{$\xrightarrow{\hspace*{.5cm}}$ Pretraining on Cityscapes} & \multicolumn{13}{l}{$\xrightarrow{\hspace*{1.8cm}}$ Adaptation on KITTI-360 $\xrightarrow{\hspace*{1.6cm}}$ Adaptation on SemKITTI-DVPS $\xrightarrow{\hspace*{.4cm}}$} \\
        \\[-1.5ex]
        Cityscapes & 72.87 & 49.19 & 77.45 & 60.40 & 10.16 & 0.19 & 72.90 & 49.76 & 77.49 & 61.22 & 11.38 & 0.21 & 72.42 & 48.74 & 77.08 & 60.20 & 10.65 & 0.21 \\
        KITTI-360 seq. 10 & 45.74 & 30.62 & 69.56 & 39.49 & 7.90 & 0.33 & 49.91 & 31.91 & 70.68 & 40.95 & 5.57 & 0.15 & 49.26 & 32.32 & 64.08 & 40.95 & 5.23 & 0.15 \\
        SemKITTI-DVPS seq. 08 & 51.95 & 45.24 & 76.07 & 57.20 & 6.17 & 0.34  & 49.48 & 43.26 & 74.24 & 57.26 & 5.60 & 0.21 & 53.70 & 46.50 & 76.53 & 59.43 & 4.32 & 0.16 \\
        \arrayrulecolor{gray}
        \midrule
        \arrayrulecolor{black}
        & \multicolumn{5}{r}{$\xrightarrow{\hspace*{.5cm}}$ Pretraining on Cityscapes} & \multicolumn{13}{l}{$\xrightarrow{\hspace*{1.5cm}}$ Adaptation on SemKITTI-DVPS $\xrightarrow{\hspace*{1.5cm}}$ Adaptation on KITTI-360 $\xrightarrow{\hspace*{.8cm}}$} \\
        \\[-1.5ex]
        Cityscapes & 72.87 & 49.19 & 77.45 & 60.40 & 10.16 & 0.19 & 72.75 & 49.01 & 77.36 & 60.35 & 10.82 & 0.22 & 72.51 & 48.87 & 76.98 & 60.28 & 11.41 & 0.21 \\
        KITTI-360 seq. 10 & 45.74 & 30.62 & 69.56 & 39.49 & 7.90 & 0.33 & 49.26 & 31.66 & 70.26 & 41.40 & 6.30 & 0.17 & 50.05 & 31.92 & 70.50 & 41.48 & 5.47 & 0.16 \\
        SemKITTI-DVPS seq. 08 & 51.95 & 45.24 & 76.07 & 57.20 & 6.17 & 0.34 & 52.31 & 44.29 & 75.58 & 56.87 & 4.56 & 0.16 & 53.83 & 47.29 & 76.55 & 60.01 & 4.25 & 0.16 \\
        \bottomrule
    \end{tabular}
    \footnotesize
    \net is continually applied to three domains using Cityscapes as the initial source domain and then adapting to KITTI-360 and SemKITTI-DVPS. The listed numbers on the target domains are based on protocol 2.
\end{threeparttable}
\vspace*{-.3cm}
\end{table*}

\subsection{Continual Adaptation}

Finally, we evaluate the performance of \net in the context of multi-domain adaptation, i.e., $\mathcal{S} \to \mathcal{T}_1 \to \mathcal{T}_2$. In particular, we first adapt to sequence 10 of KITTI-360 followed by sequence 08 of SemKITTI-DVPS, then we invert the adaptation order. To analyze forward and backward transfer as defined for continual learning~\cite{lopez2017gradient}, we compute the metrics on the \textit{val} split of the source and the \textit{adapt} parts of the respective target domains. We report the results in \reftab{tab:continual-learning}. Note that we use $\alpha_{\mathcal{S} \shortto \mathcal{T}_1} = 0.9$ and $\alpha_{\mathcal{T}_1 \shortto \mathcal{T}_2} = 0.7$ for updating the EMA model according to \refeqn{eqn:ema} since the network should adapt more strongly when deployed to $\mathcal{T}_2$ due to the larger amount of previously seen data. As shown in the first row of both adaptation orders, \net is able to mitigate catastrophic forgetting with respect to $\mathcal{S}$ maintaining its performance.
We make a similar observation when re-evaluating $\mathcal{T}_1$ after the second adaptation step to $\mathcal{T}_2$. In particular, \net achieves positive backward transfer on SemKITTI-DVPS when adapting to KITTI-360.
On the same adaptation order, we observe positive forward transfer for KITTI-360, i.e., the performance increases although \net was only adapted to SemKITTI-DVPS.

\begin{figure}[t]
    \centering
    \includegraphics[width=\linewidth]{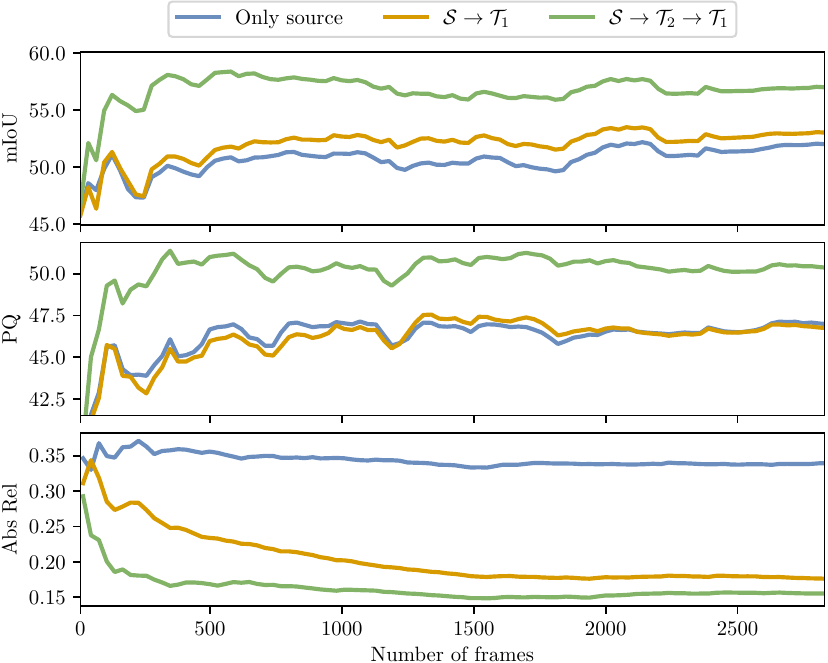}
    \vspace{-.65cm}
    \caption{Evolution of performance metrics on SemKITTI-DVPS sequence~08 during adaptation (protocol 1). The metrics are averaged until the given frame number. The target domains $\mathcal{T}_1$ and $\mathcal{T}_2$ refer to \mbox{SemKITTI-DVPS} and \mbox{KITTI-360}, respectively. It can be seen that there is positive forward transfer when first adapting on $\mathcal{T}_2$.}
    \label{fig:continual-adaptation}
    \vspace*{-.5cm}
\end{figure}

In \reffig{fig:continual-adaptation}, we illustrate the evolution of the performance metrics on SemKITTI-DVPS sequence 08 during adaptation (protocol 1). We compare the error without adaptation to directly adapting to SemKITTI-DVPS versus first adapting to KITTI-360. For both semantic segmentation and depth estimation, it can be clearly observed that the performance improves if more images have been seen. Additionally, adapting first to KITTI-360 results in a large performance increase for both semantic and panoptic segmentation. We account this to the fact that KITTI-360 sequence 10 leads to strongly improved performance, shown in \reftab{tab:continual-learning}, that can be transferred to the SemKITTI-DVPS domain.

\section{Conclusion}
In this paper, we present \net as the first approach for online continual learning for joint monocular depth estimation and panoptic segmentation. \net enables the vision system of a robotic platform to continually enhance its performance in an online fashion.
In particular, we propose a new cross-domain mixing strategy to adapt panoptic segmentation combining annotated source data with unlabeled images from a target domain. 
To mitigate catastrophic forgetting, \net leverages experience replay using a buffer composed of source and target samples.
We explicitly address the limited storage capacity of robotic platforms by setting a fixed size for the replay buffer. To ensure distinct replay samples, we use rare class sampling on the source set and employ image-based diversity sampling when updating the target buffer.
Using extensive evaluations, we demonstrate that \net outperforms competitive baselines while avoiding catastrophic forgetting in the online continual learning setting.
Future work will explore cross-task synergies and the use of pretext tasks for domain adaptation.

\section*{Acknowledgment}
This work was partly funded by the European Union’s Horizon 2020 research and innovation program under grant agreement No 871449-OpenDR and the Bundesministerium für Bildung und Forschung (BMBF) under grant agreement No FKZ 16ME0027.


\bibliographystyle{plainnat}
\bibliography{references}

\end{document}